# Automated Diagram Generation to Build Understanding and Usability
By William Schoenberg

**Abstract**
Causal loop and stock and flow diagrams are broadly used in System Dynamics because they help organize relationships and convey meaning. Using the analytical work of Schoenberg (2019) to select what to include in a compressed model, this paper demonstrates how that information can be clearly presented in an automatically generated causal loop diagram. The diagrams are generated using tools developed by people working in graph theory and the generated diagrams are clear and aesthetically pleasing. This approach can also be built upon to generate stock and flow diagrams. Automated stock and flow diagram generation opens the door to representing models developed using only equations, regardless or origin, in a clear and easy to understand way. Because models can be large, the application of grouping techniques, again developed for graph theory, can help structure the resulting diagrams in the most usable form. This paper describes the algorithms developed for automated diagram generation and shows a number of examples of their uses in large models. The application of these techniques to existing, but inaccessible, equation-based models can help broaden the knowledge base for System Dynamics modeling. The techniques can also be used to improve layout in all, or part, of existing models with diagrammatic informtion.

**Introduction**
The visual depiction of the structure of a model is very important to the ability of a model to be useful (Sterman, 2000).  Generally, model structure is consumed in the form of Stock Flow Diagrams (SFDs) or Causal Loop Diagrams (CLDs) as opposed to a direct reading of the equations with no diagrammatic representation.  Algorithmic generation of model diagrams is a critical component to model translation, and arguably to improving model construction and understanding in general.  Human driven model diagram generation is well understood to be critically important to the model understanding and construction process, depending heavily upon the quality of the layout, its high-level organization, its ability to emphasize loops, make clear key structures, and properly organize content (Richardson, 1986).  Likewise, algorithmic development of model diagrams is also key to improving understanding and the construction of models.  At the very least, algorithmic diagram construction represents the ability to save a significant human effort when it comes to model presentation.

Before the mid to late 1980's, and the rise of graphical software to build models, models were built purely as systems of equations represented by the syntax of the programming language, typically DYNAMO, that they were constructed in.  For models built in that era to be used now-a-days, they have to be translated into modern modeling languages which typically support and sometimes require a diagrammatic representation.  Automated translation systems have been developed for DYNAMO to XMILE (Ward et. al, 2015) which handle the equations, but until now, the process of creating the SFDs representing that system structure has had to be done by hand.  One of the best such examples of an important model locked into the era of non-diagrammatic representation is Forrester's 'National Model' (3348 equations).  The last version of the model worked on by Forrester is a Vensim mdl model, text-only formatted equations, lacking any visual elements. One of the major challenges to its continued development and use has been that it lacks an SFD.

There are two major components to algorithmic diagram generation. The first, and most well studied is the process of laying out a directed graph of nodes and edges, or as system dynamists call it, the process of drawing and arranging variables and links. This challenge has been very well solved in the past, especially by Schoenberg (2009, 2019) who has built on decades of research into algorithmic graph drawing, applying that research to directed cyclic graphs of which system dynamics models are just one example. For SFDs, a sub-challenge unique to the SD field, is the process of laying out and drawing stock and flow chains, since these symbols have a unique set of constraints not typically encountered by those in the graph drawing field.

The second major component to the challenge of diagrammatically representing large models, in particular SFDs, is clustering. Specifically, where to draw module boundaries, and how to arrange model content such that like concepts are grouped together while maintaining readable diagrams of manageable levels of cognitive complexity. Typically, the clustering problem is iteratively solved by the original modeler who groups equations manually into organizational units which are based around the key conceptual areas covered by the model. Recreating this very human process in an automated fashion is a challenge which must be overcome by any algorithm which claims to generate useful SFDs for large, or even moderately complex models. It is not enough to just layout main-chains and connect up the symbols representing the equations. In addition to the layout problems mentioned above, model equations must be partitioned into logical groupings (modules) in order to maintain some control on the cognitive complexity of the resulting model diagram. While the idea of clustering is required for SFDs, for CLDs modelers have more freedom when it comes to controlling the cognitive complexity of their diagrams, because a CLD does not have to depict each and every equation within the model. While a modeler could for instance apply any clustering solution developed for SFDs to CLDs, with a CLD modelers are free to follow the recommendations of Schoenberg, 2019 to algorithmically simplify and aggregate the model's depiction using the link and loop threshold parameters.

This paper demonstrates the algorithms and techniques necessary to autogenerate SFDs and CLDs for large models based upon the network structure encoded by the equations. Building upon the work of Schoenberg (2019), this paper demonstrates how to expand the scope of that technology to models such as Forrester's 'National Model', as well as to other forms of diagrammatic representations, including SFDs. In addition, this paper uses existing research in the machine learning field for community detection and modularity to present a novel solution to the modularity problem introduced above. This work allows for automated improvements in system representation and presentation in production quality software (Stella v2.0). Specifically, it produces well-formed and accurate representations based upon the structure of the system. It allows for the wholesale creation of diagrams for legacy models that lack them, or for the improvement in form and organization of diagrams that already exist.

**Review: Techniques for machine generation of network diagrams**

Schoenberg (2019) best demonstrates the state of the art in algorithmic diagram generation in directed cyclic graphs.  The basis for this work, and that work is a Kamada Kawai force directed layout algorithm.  A force directed layout algorithm solves the problem of the placement of nodes in 2D space, such that symmetry is generated, and edge length is approximately equal, by running a physics simulation of weights connected by springs and minimizing the total energy of the system. The first force directed layout technique used steel rings to represent each node and then connected those rings using logarithmic springs (Eades, 1984). In this version of the algorithm, attractive forces were only calculated between neighbors, and repulsive forces were calculated between all node pairs (Eades, 1984). This process ensured that neighbors were always close by but limited the scope of the N-squared problem.

The next evolution in the force directed approach was to introduce the concept of an ideal distance between every node pair based on the shortest path between each node pair, and to use Hooke's Law, meaning real world realistic linear springs (Kamada and Kawai, 1989). The Kamada Kawai approach solved partial differential equations based on Hooke's Law to optimize layout applying all forces between all node pairs in an iterative fashion (Kamada and Kawai, 1989). A gradient descent optimization process used to terminate the simulation when a global minimum in the energy state of all the springs was found (Kamada and Kawai, 1989).

Development of Graphviz, an open source toolkit for solving these graph generation problems took place in parallel to these developments. Graphviz contains many different automated layout mechanisms, but the mechanism most relevant to SFD and CLD generation is called neato, which is based on the work of Eades, Kamada and Kawai among others. The layout algorithm used in neato that we are interested in, is derived from the Kamada Kawai algorithm. It assumes there is a linear spring between every pair of nodes, each with an ideal length (Gansner, 2014). The ideal distance between each node pair is the result of a function computed for each pair; the ideal length function relevant to SD diagram generation uses the shortest path between the two nodes to determine the ideal distance between these nodes, although many other choices are offered. Neato is able to turn a static text file with a description of the graph quickly into a 2D diagram quickly (North, 2004).

As a part of the development of LoopX, Schoenberg (2019) solved the problem of how to curve the straight edges produced normally in standard force directed layout algorithms.  Schoenberg applied a simple heuristic which states that the center of the cycle which forms the arc that the edge will follow, must be the average center of the nodes which form the shortest feedback loop with length greater than 2 that the edge is a member of. The two-node exception is handled separately within the neato codebase and produces paired directed edges that do not over-emphasize the cycle, - producing elongated ellipse structures that cover an area relative to the number of nodes. This edge curving heuristic relies upon the attributes of force directed graphs which place nodes that are related closest together. This heuristic produces loops that look circular except for in degenerate cases where the force directed layout fails to produce good local clusters of neighbors in the graph space, and the shortest feedback loops are relatively far flung in the 2D space.

**Review: Community detection algorithms, detecting subcomponents in large network graphs**

The problem of automated modularity of models is most akin to the problems of community detection and modularity as conceived by those who study large scale networks. These fields offer a rich literature to study. They define the problem from the following perspective: how can the nodes of a network be grouped such that there is a higher density of edges within the clusters then between the clusters (Girvan & Newman, 2002; Clauset et. al, 2004; Yang et. al, 2016). From the perspective of a SD model, the nodes are variables (equations), the edges are the relationships between the variables as specified by the equations. Putting the community detection problem into the language of models and variables, yields the following statement: how can the variables of a model be grouped such that there is a higher density of links within each module then there is between the modules. This is exactly the same problem human modelers face when they organize and partition their variables into modules.

Research into community detection started in 2002 with the seminal work by Girvan and Newman, '*Community structure in social and biological networks*'. In this paper the authors discuss and define the concept of a community and then propose a hierarchical algorithm to identify communities within a network. Their algorithm follows four steps, producing a dendrogram, where the algorithm selects the level of the dendrogram which exhibits the smallest ratio between the density of edges between the clusters to the density of edges within the clusters. This algorithm runs in exponential time.

1. The betweenness (centrality) of all existing edges in the network is calculated
2. The edge(s) with the highest betweenness are removed
3. The betweenness of all edges affected by the removal is recalculated.
4. Steps 2 and 3 are repeated until no edges remain.

This algorithm typically serves as the benchmark by which all other algorithms are judged, but relatively speaking is considered slow (Yang et. al, 2016).

The next major improvement in community detection algorithms came along in 2004, with the work of Clauset, Newman, and Moore who developed a hierarchical agglomerative algorithm which runs in near linear time for sparse and hierarchical networks, or logarithmic time in the worst case, without sacrificing quality on the communities discovered. Sparse hierarchical networks are those where not all nodes are connected, and there is a clear hierarchy of the relationships. Typically, system dynamics models fall into this category. The practical effect of their work in this space is that system dynamics models of incredible complexity, meaning hundreds of thousands of variables with millions of links can be quickly and efficiently grouped into logical modules, while minimizing the ratio between the density of links between the modules relative to the density of links within the modules.

The Clauset Newman Moore algorithm follows the following steps:

1. For a partition of a given network, define the modularity, $Q$, of the partition to be the ratio of the number of edges within each community to the number of edges between each community, subtracted from the ratio you'd expect from a completely random partition.
2. Given two communities $i$ and $j$ of a partition, define $\Delta Q_{ij}$ to be how much the modularity of the partition would change if communities $i$ and $j$ were merged.
3. Apply a simple greedy algorithm: start with every node in its own community. Calculate $\Delta Q_{ij}$ for every pair of communities. Whichever two communities $i$ and $j$ have the largest $\Delta Q_{ij}$, merge those two. Repeat step 3 until maximum modularity is reached when all $\Delta Q_{ij}$ turn negative.

There are other approaches to solving this problem, and research is still on-going to the present day, but for the purposes of determining modules in system dynamics models, knowledge of Girvan Newman and Clauset Newman Moore is all that is necessary. Both of these clustering algorithms have been implemented in the popular open source toolkit, SNAP and these reference implementations are used for all work presented in this paper (Leskovec & Sosič, 2016).

**Solutions to the challenges of generating SFDs using force directed layout algorithms**
Before discussing modularity, and generating diagrams for large models, first it must be understood how to adapt force directed layout algorithms to generating simple SFDs. The challenge here are the additional rules introduced by the inclusion of stocks and flows, specifically that flows can only contain right angles. The demonstrated solution to this problem turns the SFD diagram layout into a two-stage process. The first stage of the process is to find and layout all stock flow main chains which are then treated as single nodes, equivalent to single variables for the second stage of the layout process which uses neato to perform the actual layout. The agglomerated nodes are referred to as, main chain nodes.

**Finding and laying out main chains**
Identifying main chains from the equations of a model is a simple process. All stocks are iterated, and if any flow of the stock is attached to any other stock (upstream or downstream) then we know that stock is a member of a main chain and must therefore be agglomerated into a main chain node. The agglomeration process walks the flows into and out of the candidate stock, storing all stocks which have been explored until there are no more flows in the sub-network to explore. Only a single main chain node is generated for each main chain, meaning during the iteration of stocks, each stock is checked to make sure it isn't a member of an already discovered main chain. Individual stocks with any arbitrary number of flows are also agglomerated into a main chain node of their own containing just the single stock and its flows.

Main chain nodes must first perform their own internal layout process, each in their own unique 2d coordinate space before the layout of the larger diagram can take place. Start with the stock most central to the network of the main chain (only using flows for edges), this stock will be the one which has the shortest average path-length to all other stocks in the network. This stock is marked as visited, and therefore ineligible for future layout, and is placed at coordinate 0,0 of the unique coordinate space which represents the layout of the main chain being operated on.  This central stock is the first candidate stock for the following algorithm.  The number of inflows (then outflows) is determined and multiplied by the height of a stock plus a value for spacing.  The $x$ coordinate for any upstream stock is set to a fixed width to the left of the candidate stock, for downstream stocks, it is offset by the same fixed width to the right of the candidate stock.  To determine the starting $y$ coordinate of the first upstream or downstream stock, take the height determined in the previous step, and center it relative to the candidate stock whose upstream or downstream stocks are being laid out.  Then begin the process of iterating through each flow in the set, and when a stock which has not already been laid out is encountered place it in the appropriate space allocated when doing the height calculation and mark it as visited.  Make sure to leave blank slots allocated for stocks which do not actually exist.  This ensures there is appropriate room to layout all flows into the candidate stock.  To finish, flows must be drawn connecting the stocks using 4 points arranged in a 3 segment 'S' pattern.  Flows leave their upstream stocks from the right side and enter their downstream stocks from the left.  For flows which do not connect two stocks or flows which are completely straight a single horizontal segment is used instead.  Flows are then properly spaced on each inbound or outbound side of the stock to prevent overlap. This process ensures an organized and logical layout of any arbitrary stock flow main chain in 2d space.

**Laying out diagrammatic elements of SFDs or CLDs using neato**
The second major step in the layout of an SFD or CLD is walking the nodes and edges of a model and running a Kamada Kawai force directed graph algorithm on the network to get a final diagram.  This section of the paper will describe how to do that with highly dis-similar node types (main chain nodes vs. single variable nodes).  The main problem here are the issues related to node overlap.  This problem is exacerbated by the construction of main chain nodes for SFD diagram generation, because those nodes can become particularly large in the case of complex main chains.  When dealing with main chain nodes it becomes very important to specifically keep track of, and re-assign all the edges into and out of any element (stock or flow) in the main chain node to the main chain node itself, lest links between variables be lost for the purposes of the layout algorithm.  After the diagram has been generated with main chain nodes, re-assign the edges back to the proper elements within the main chain nodes.

To solve issues with overlap, use the built-in overlap removal algorithms within neato.  For the overlap algorithms within neato to work, size information must be given for each node.  For each single variable node, specify the size of the symbol, plus the variable name, plus an arbitrary size margin.  For main chain nodes, take the union of all space occupied by all stocks and flows, plus the same arbitrary margin from single nodes.

The two most useful overlap removal algorithms which ship with neato by default are Voronoi and VPSC (PRISM can be better but is not included in the base install and is therefore ignored for the purposes of this discussion).  Figure 1 demonstrates the importance of overlap removal, and the differences between the techniques.  The Voronoi technique is iterative, on each iteration a bounded Voronoi diagram of the node center points is computed, and each node is moved to the center of its Voronoi cell. This is repeated until all overlaps are eliminated (Gansner & North, 1998).  This is potentially useful for larger diagrams because it tends not to affect the overall aspect ratio, but its downside is that it tends to over expand the diagram, pushing nodes towards the bounds of the diagram, creating too much margin between neighboring elements.  The VPSC technique is a quadratic programming algorithm, based on setting up pairwise node constraints to remove overlaps, and sets up two different problems for the vertical and horizontal orientation (Gansner & Hu, 2009).  VPSC is generally most useful for system dynamics diagrams, because it works well to remove overlaps in diagrams of ~100 variables or less without introducing excess margin, or heavily adjusting the diagram aspect ratio.  In diagrams of larger sizes, VPSC tends to produce elongated diagrams in the vertical orientation.

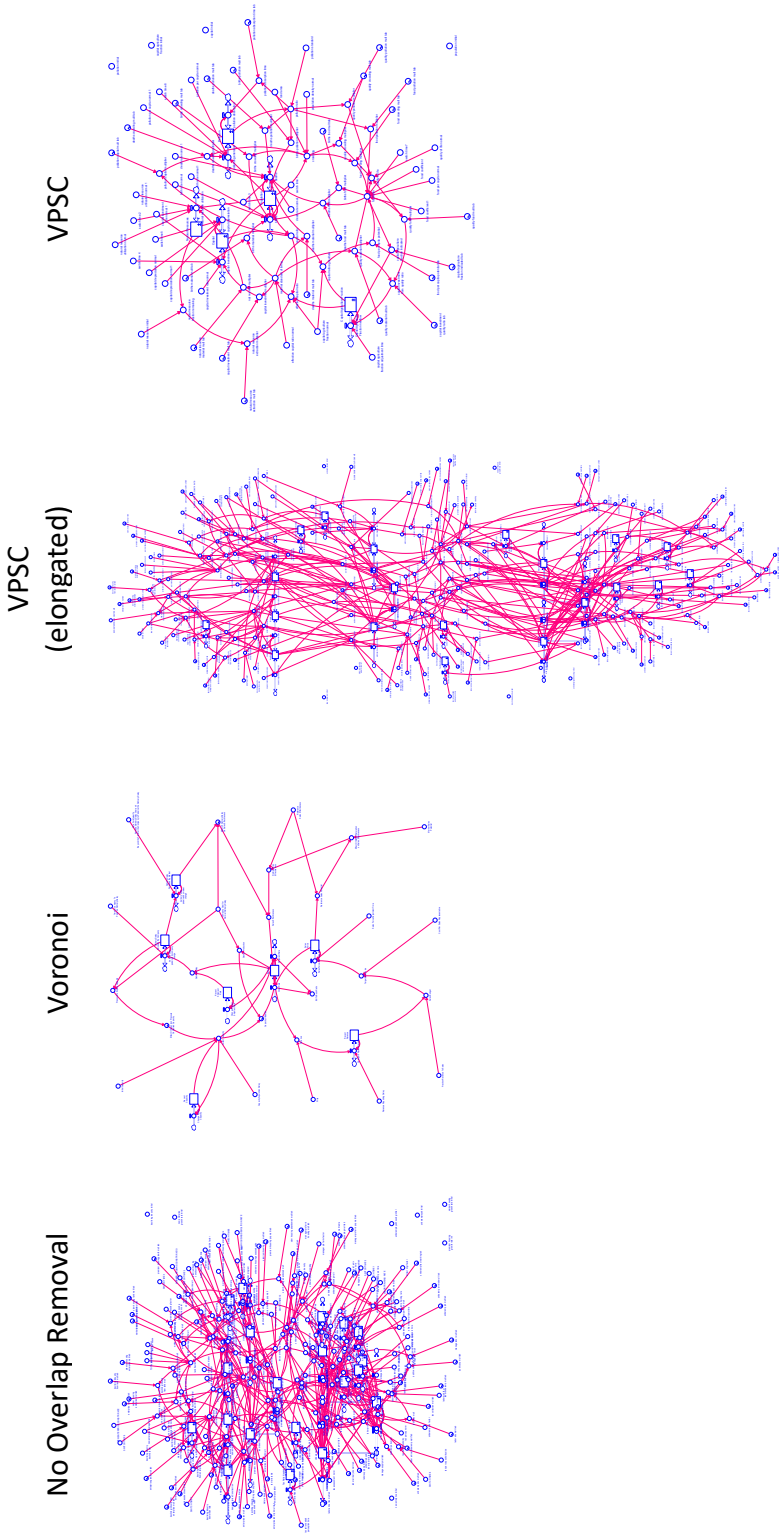

*Figure 1: Demonstration of node overlap removal techniques. No overlap removal and VPSC (elongated) were produced using Urban Dynamics (319 symbols). Voronoi was produced using Market Growth (42 symbols) and VPSC was produced using World2 (100 symbols).*

**Applying community detection algorithms to SD models**
Figure 1 demonstrates the issues with attempting to generate diagrams for large models without grouping content into modules.  The diagrams which are generated are too large to effectively read, and the incidence of degenerate diagrams rises as more and more complexity is packed in.  The solution to this problem ensures a consistent diagram size by recursively applying Clauset Newman Moore community detection to the network of model equations with a threshold of 75 or more variables triggering a clustering of any individual module into sub-modules.  This breaking point provides an appropriate balance between diagram complexity and number of modules, ensuring that each generated module diagram is both readable, and well related.

The application of any clustering technique to a model is a fairly straight forward process.  This paper describes the process using the SNAP implementation of these algorithms, but these steps are robust to the implementation of any Girvan Newman, or Clauset Newman Moore clustering algorithm.  For the case of all diagrams presented in this paper, SNAP's Clauset Newman Moore implementation was used.  Like in diagram layout, for modularity, the first step is to identify and group together all stocks and flows into main chain nodes.  As described above, make sure to redirect any links incoming or outgoing from the stocks or the flows of a main chain to the main chain node itself.  The SNAP implementation of these algorithms operates on two files, the first is the input file, which specifies a list of directed edges between nodes, the second is the output file which clusters the nodes, the third argument to the program controls which clustering algorithm to use.  To generate the input file, walk all the nodes (variables, and agglomerated main chains) and assign each a unique id.  Then from each node, write out either all of the incoming connections *or* outgoing connections (not both) using the unique ids.  After running SNAP's community program, the output file will then list the set of unique ids which belongs to each cluster, never duplicating unique ids across clusters, and because the main chains are represented by a single node there is no need to worry about a main chain being split across modules.

**Application of these techniques to World2 and The National Model**
To demonstrate the efficacy of these algorithms let's analyze the generated diagrams for World2 and Forresters' National Model.

**World2**
World2, contains 100 symbols originally organized into 3 modules: 'Population & Food', 'Capital & Quality Life' and 'Pollution & Resources'.  The generated top level diagram for this model, Figure 2, contains 8 modules, which have been labeled by hand after the diagram generation process (the only non-machine input into these diagrams).  Figure 3, demonstrates the main-chain node layout process, as applied to the Population module of the autogenerated diagram.

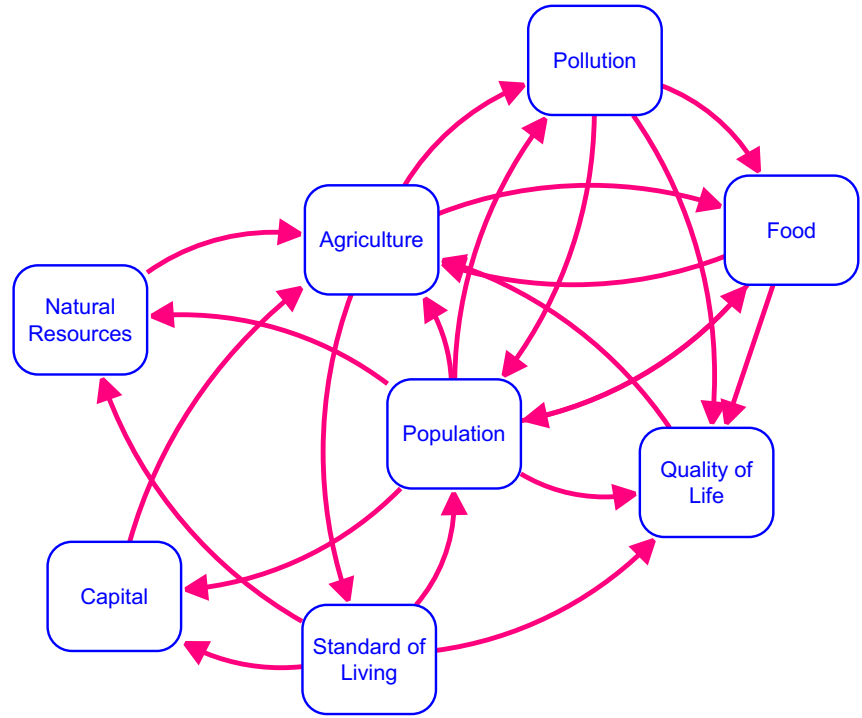

*Figure 2: Autogenerated top-level diagram for World2*

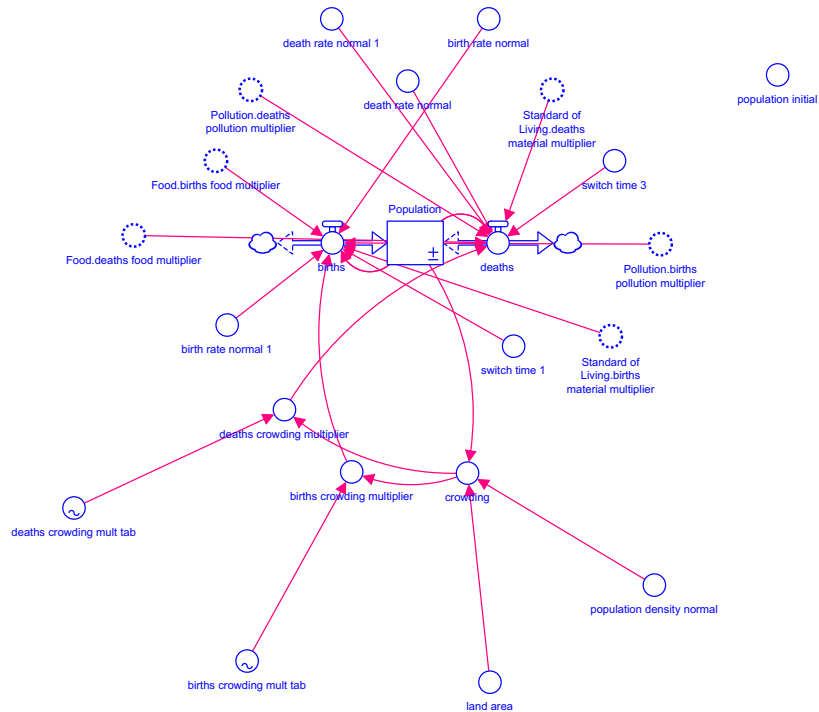

*Figure 3: Autogenerated diagram of the Population module of the World2 model*

Figure 3 demonstrates one of the major weaknesses of the agglomerative main-chain approach as discussed above. Because 'Population', 'births', and 'deaths' are laid out as a single node with no distinction between the sub-elements, the force directed graph is not able to ideally place all of the direct causes and uses from it, because it does not know the location of the actual source or target for those connectors. If it did, this would correct the problem with 'deaths crowding multiplier' being to the left (under 'births') rather than to the right, (under 'deaths'), which would have then changed the position of 'deaths crowding mult tab'. The same is true for the crossing connectors above 'Population'. Overall though, the shows strong symmetry, and emphasis on the feedback loops, vs the non-dynamic relationships and is easily correctable by a human if necessary.

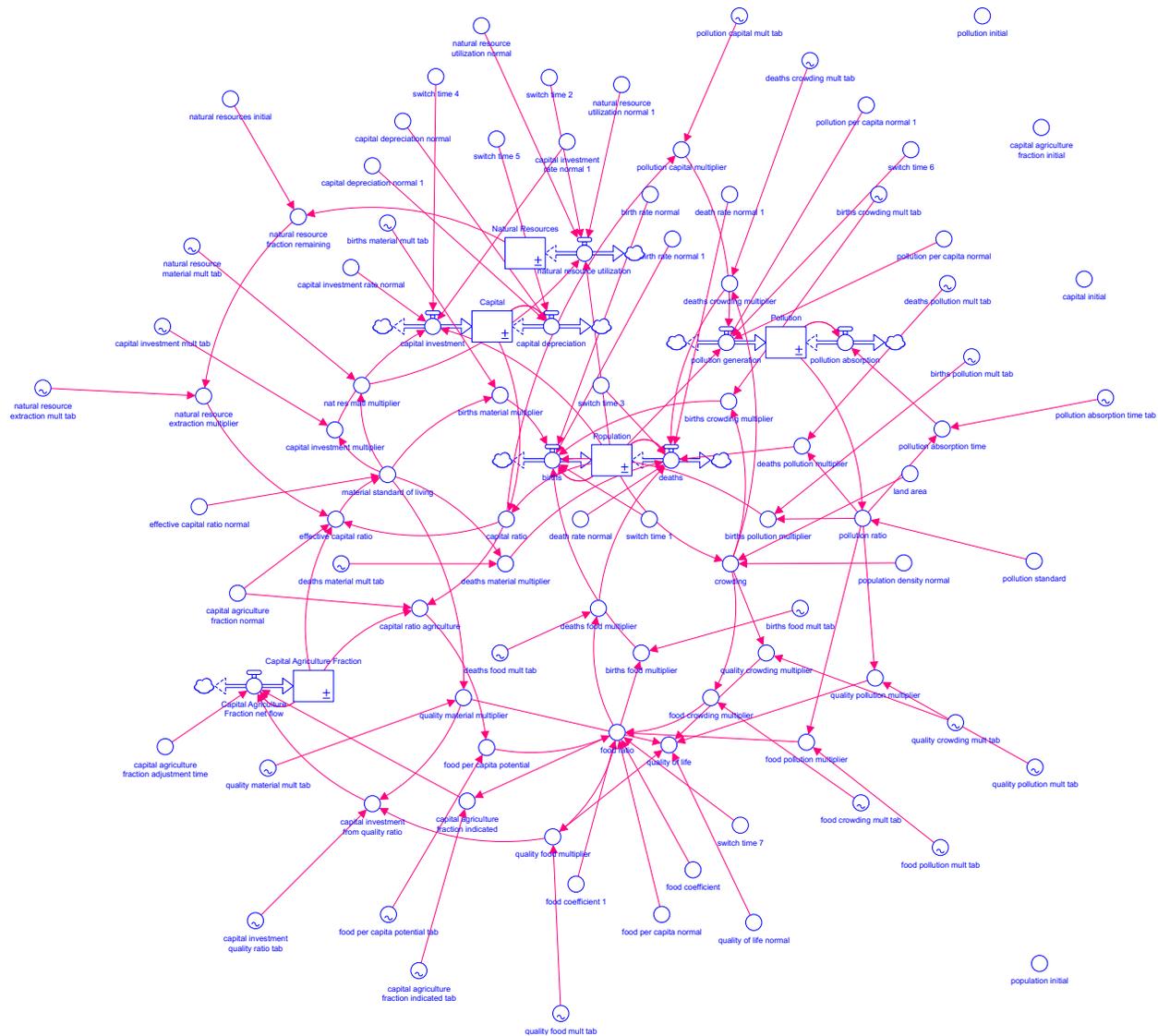

*Figure 4: Autogenerated diagram for World2 without performing any modularity.*

Figure 4 demonstrates the utility of the modularity process by showing what World2 looks like laid out as a single module. It is easier to see the complex network of high-level relationships between the key model areas in Figure 2, then Figure 4. This diagram does show the power of the force directed graph algorithm to do good layout in a very difficult situation by maintaining strong symmetry and emphasizing the major feedback loops between the stocks. Realistically though, with the primary goal of this process being to improve the understanding and construction of models, the diagram in Figure 4, is much less successful then the diagrams shown in Figure 2 and Figure 3 reinforcing the importance of clustering.

**The National Model**
The National Model developed by Forrester contains 2,446 symbols.  Up until the writing of this paper, the model was last worked on as a mdl file (text format equations) containing no diagram information.  Forrester did produce a hierarchy of equations which were usable as hints for performing the modularity process.  This model is so large that generating a single diagram for all symbols would be a fool's errand, therefore this paper compares and contrasts the generated diagrams using Forrester's hierarchy and a Clauset Newman Moore community detection algorithm.

Figure 5 shows the modularity using Forrester's hints.  In it there are 14 top level modules, and 88 total modules.  The labor sector module in Figure 5 is a reasonable example of the complexity of the generated diagrams within each of the modules.  Most modules contain some structure and some sub-modules.  The model in this form is now eminently usable, and can continue to be constructed, and tested.  Forrester's hints produce modules which tend to minimize the number of connections at the higher levels where possible.  When taken in conjunction with Figure 6, which demonstrates the output of a Clauset Newman Moore modularity, it is easily seen just how interconnected this model is at its higher levels.  The Clauset Newman Moore modularity version contains 112 modules, with 21 linked modules at the top level.  The unlinked modules and variables at the top level of the Clauset Newman Moore version are testing variables which Forrester lumped together in the Parameter Values module.  None of those variables are material to the model in its current form, but none the less, this demonstrates a weakness of any modularity algorithm, because it cannot know intent for disconnected structures.

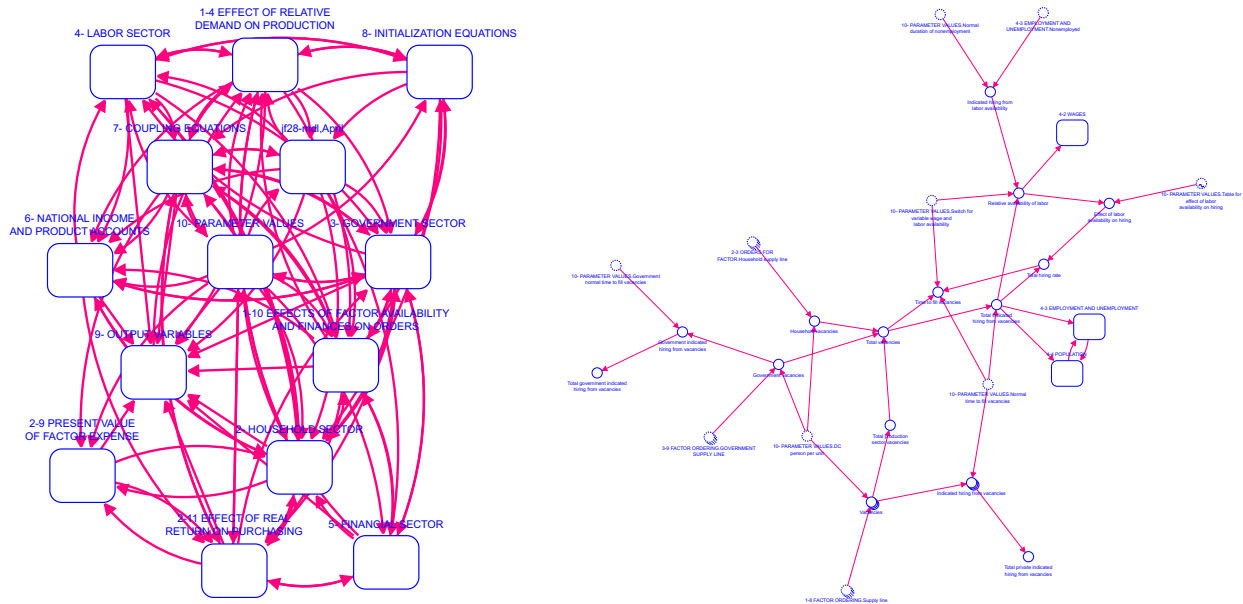

*Figure 5: On the left is the top-level view of modules from National Model using Forrester's hints. On the right is Labor Sector from Forrester's National Model using Forrester's modularity hints. The labor sector is representative of the typical complexity of any of the generated modules*

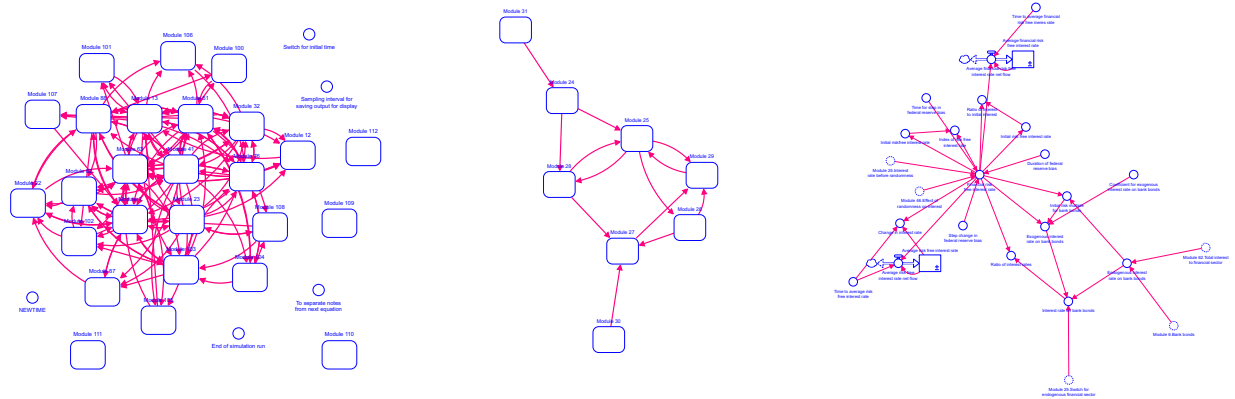

*Figure 6: On the left is Forrester's National Model, top level diagram using Clauset Newman Moore modularity. In the center is Module 23 (Finances and Risk) which is solely built up of other modules, one of which is show on the far right, Module 26, which is concerned with interest rates.*

Overall, both approaches to creating SFDs for the National Model have been successful. Both produce similar numbers of modules, and diagrams of comparable levels of complexity. Both represent a major success on the primary goal of improving the understanding and further construction of this model.

**Conclusions**

This paper has demonstrated how to algorithmically produce structure diagrams for large models such as Forrester's National Model quickly and efficiently. It has demonstrated techniques for algorithmically determining the divisions between modules, and a process for laying out main chains as a part of the full network of model equations. These algorithms represent a major success because they allow for the automated translation of models from languages like DYNAMO which lack a diagrammatic display, into modern software platforms which require them. These algorithms also offer the opportunity to let the machine do most of the laborious, grinding, work of model presentation. The diagrams generated by these processes tend to have the attributes of well-developed diagrams, including symmetry with emphasis on the loops. At the very least, this approach demonstrates a strong starting point after which a human modeler can take over and do any tuning necessary to perfect the presentation of model logic. Future work in this area needs to focus on improving the main chain node concept, giving the force directed graph algorithm the ability to know the location of sub-elements, while forcing it to deal with the main chain as a single unit. Additionally, it would be an improvement to develop an algorithm to automatically name the generated modules. At the very least this work helps to improve the presentation and understanding of model logic in general.

**References:**
1. Clauset, A., Newman, M. E., & Moore, C. (2004). Finding community structure in very large networks. *Physical review E*, *70*(6), 066111.
2. Eades, P. (1984). A heuristic for graph drawing. *Congressus numerantium*, *42*, 149-160.
3. Gansner, E. R., & North, S. C. (1998, August). Improved force-directed layouts. In *International Symposium on Graph Drawing* (pp. 364-373). Springer, Berlin, Heidelberg.
4. Gansner, E. R., & Hu, Y. (2010). Efficient, proximity-preserving node overlap removal. In *Journal of Graph Algorithms and Applications* (pp. 53-74).
5. Gansner, E. R. (2014). Using Graphviz as a Library (cgraph version). *published online August 21.*
6. Girvan, M., & Newman, M. E. (2002). Community structure in social and biological networks. *Proceedings of the national academy of sciences*, *99*(12), 7821-7826.
7. Kamada, T., & Kawai, S. (1989). An algorithm for drawing general undirected graphs. *Information processing letters*, *31*(1), 7-15.
8. Leskovec, J., & Sosič, R. (2016). Snap: A general-purpose network analysis and graph-mining library. *ACM Transactions on Intelligent Systems and Technology (TIST)*, *8*(1), 1.
9. North, S. C. (2004). NEATO user's guide. *Murray Hill, NJ: AT&T Bell Laboratories*.
10. Richardson, G. P. (1986). Problems with causal-loop diagrams. *System dynamics review*, *2*(2), 158-170.
11. Schoenberg, W. (2009). The Effectiveness of Force Directed Graphs vs. Causal Loop Diagrams: An experimental study. In *The 27th International Conference of the System Dynamics Society*.
12. Schoenberg, W. (2019). LoopX: Visualizing and understanding the origins of dynamic model behavior. *arXiv preprint arXiv:1909.01138*.
13. Sterman JD. 2000. *Business Dynamics: Systems Thinking and Modeling for a Complex World.*Irwin/McGraw-Hill, Boston.
14. Ward, Robert, James Houghton, and Ivan A. Perl. "SDXchange: stand-alone translators to enable XMILE model adaptation, transportation, and exchange." *System Dynamics Review* 31.1-2 (2015): 86-95.

15. Yang, Z., Algesheimer, R., & Tessone, C. J. (2016). A comparative analysis of community detection algorithms on artificial networks. *Scientific reports*, *6*, 30750.